\documentclass[letterpaper]{article} 
\usepackage{aaai2026}  
\usepackage{times}  
\usepackage{helvet}  
\usepackage{courier}  
\usepackage[hyphens]{url}  
\usepackage{graphicx} 
\usepackage{natbib}  
\usepackage{caption} 
\usepackage{algorithm}
\usepackage{algorithmic}

\usepackage{enumitem}
\usepackage{amsmath}
\usepackage{amsfonts}
\usepackage{booktabs}
\usepackage[capitalize]{cleveref}
\usepackage{nameref}

\crefname{section}{Sec.}{Secs.}
\crefname{section}{Section}{Sections}
\crefname{table}{Table}{Tables}
\crefname{table}{Tab.}{Tabs.}

\usepackage{multirow}
\usepackage[table]{xcolor}

\def\model{TGE}
\def\ds{Colored Shape Fidelity}

\makeatletter

\newcommand{\Rmnum}[1]{\textcolor{red}{\expandafter\@slowromancap\romannumeral #1@}}

\makeatother
\usepackage{newfloat}
\usepackage{listings}
\DeclareCaptionStyle{ruled}{labelfont=normalfont,labelsep=colon,strut=off} 
\floatstyle{ruled}
\newfloat{listing}{tb}{lst}{}
\floatname{listing}{Listing}
\title{Textured Geometry Evaluation: Perceptual 3D Textured Shape Metric via 3D Latent-Geometry Network}
\author{
    Tianyu Luan\textsuperscript{\rm 1}, Xuelu Feng\textsuperscript{\rm 1}, Zixin Zhu\textsuperscript{\rm 1}\thanks{Corresponding author}, Phani Nuney\textsuperscript{\rm 1}, Sheng Liu\textsuperscript{\rm 1}, Xuan Gong\textsuperscript{\rm 1 \rm 2}, David Doermann\textsuperscript{\rm 1}, Chunming Qiao\textsuperscript{\rm 1}, Junsong Yuan\textsuperscript{\rm 1}
}
\affiliations{
    \textsuperscript{\rm 1} State University of New York at Buffalo\\
    \textsuperscript{\rm 2} Harvard Medical School\\
    \{tianyulu, xuelufen, zixinzhu, phaneesw, sliu66, xuangong, doermann, qiao, jsyuan\} @buffalo.edu
}

\usepackage{bibentry}

\begin{document}

\maketitle

\begin{abstract}
Textured high‑fidelity 3D models are crucial for games, AR/VR, and film, but human-aligned evaluation methods still fall behind despite recent advances in 3D reconstruction and generation. Existing metrics, such as Chamfer Distance, often fail to align with how humans evaluate the fidelity of 3D shapes. Recent learning-based metrics attempt to improve this by relying on rendered images and 2D image quality metrics. However, these approaches face limitations due to incomplete structural coverage and sensitivity to viewpoint choices. Moreover, most methods are trained on synthetic distortions, which differ significantly from real-world distortions, resulting in a domain gap. To address these challenges, we propose a new fidelity evaluation method that is based directly on 3D meshes with texture, without relying on rendering. Our method, named \textit{Textured Geometry Evaluation} \model{}, jointly uses the geometry and color information to calculate the fidelity of the input textured mesh with comparison to a reference colored shape. To train and evaluate our metric, we design a human-annotated dataset with real-world distortions. Experiments show that \model{} outperforms rendering-based and geometry-only methods on real-world distortion dataset.
\end{abstract}

\begin{figure*}[ht]
  \centering
   \includegraphics[width=0.9\linewidth]{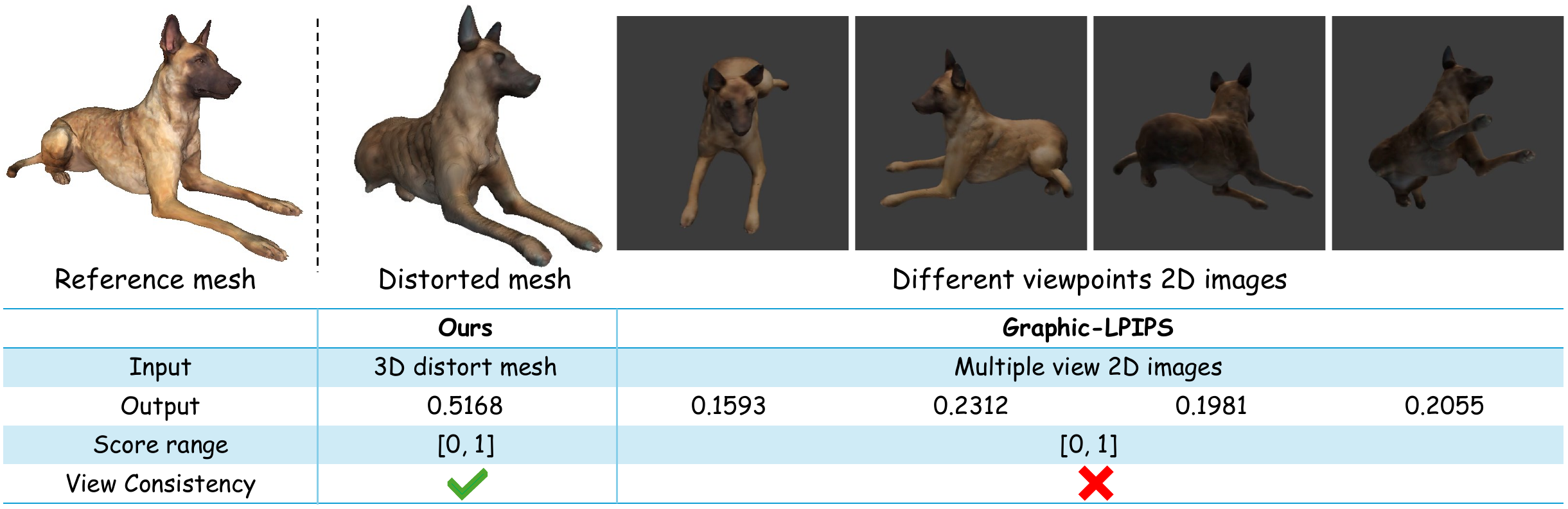}
   \caption{Comparison between previous rendering-based evaluation and our proposed 3D latent-geometry-based metric. Prior works, such as Graphic-LPIPS, rely on rendering 3D shapes into 2D images and assessing fidelity using image-based quality metrics, which makes this approach sensitive to viewpoint. Such methods produce inconsistent scores depending on rendering conditions (Right). In contrast, our method directly operates on textured 3D meshes, without relying on rendering (Left).}

   \label{fig:teaser}
\end{figure*}

\section{Introduction}
3D reconstruction and generation is a field with broad applications, covering scenarios such as video games, AR/VR, and film production. High-fidelity 3D shapes with texture are crucial in representing the spatial content in these 3D applications. Current methods focus primarily on reconstructing and generating high-fidelity 3D shapes, and have already produced many widely appreciated results. However, most of the widely used shape fidelity evaluation metrics, such as Chamfer Distance \cite{borgefors1984CD}, are still not aligning with human perception well. Therefore, evaluating the fidelity of textured 3D shapes in a human-like manner can better support the generation of high-fidelity 3D models that align with user preferences.

There has been recent progress in fidelity evaluation metrics for textured 3D shapes that are aiming to align with human perception. Those works, such as \cite{nehme2023textured} would largely rely on rendering. As shown in \cref{fig:teaser}, these methods typically render 3D shapes into 2D images and evaluate shape fidelity based on the viewpoints of the rendered image. Although such methods can leverage 2D image quality assessment techniques to evaluate 3D shapes, they have limitations. Specifically, rendered images cannot fully reflect the overall structure of a 3D shape. To comprehensively assess the fidelity of a 3D shape, multi-view and full-coverage rendering is required, which significantly increases both the computational cost and the uncertainty of evaluation with viewpoint choice. 
In addition, most existing methods (e.g., \cite{nehme2023textured}) rely on training datasets constructed from synthetic distortions. However, the distortions in synthetic data differ substantially from those encountered in real-world reconstruction and generation methods, leading to a significant domain gap. Models trained on synthetic datasets would experience substantial performance degradation when applied to real-world 3D fidelity evaluations.

To address the problems introduced by 2D-rendering-based evaluation and the domain gap caused by synthetic data, we propose a dataset constructed from real data and a 3D evaluation metric based directly on 3D meshes and textures without rendering. First, to eliminate the uncertainty introduced by viewpoint difference, our network is designed to take both 3D mesh and texture as input without reliance on viewpoints. Second, during the 3D fidelity annotation process, we rely on human subjects’ perception of fidelity, with each object evaluated by multiple subjects. To reduce subjective inconsistency caused by varying viewpoints, we continuously rotate each 3D object across multiple angles and adjust lighting directions during the evaluation. This ensures that subjects can comprehensively observe the shape under various directions and make their fidelity evaluations more consistently. Additionally, to eliminate the domain gap caused by synthetic data, all distortions in our dataset are generated by different real-world reconstruction and generation algorithms. Through such training, our model can robustly evaluate 3D shapes without being largely affected by viewpoint changes or domain gap.

We name our method \textit{Textured Geometry Evaluation} (TGE). Our network encodes both the input colored shape and the reference shape into the latent feature, and compares them in the latent space to determine the fidelity of the input shape. Specifically, we are inspired by PointNet++ \cite{qi2017pointnet++} and design a multi-stage Latent-Geometry network to extract features from colored meshes. We introduce a cross-attention mechanism to encode the color feature and fuse it with geometric features, resulting in shape representations that incorporate both texture and geometry. Using this encoding strategy on both the input and reference shapes, we are able to map the input and reference shapes into the same latent space. A fidelity comparison module is then used to compute the fidelity difference between the input and reference shapes, giving the final fidelity score for the input shape. We train and evaluate this network using our provided \ds{} dataset. With this design, the proposed fidelity shape evaluation metric does not rely on viewpoints and aligns well with human perceptual fidelity evaluation based on real-world data distribution.

Our contributions are summarized as follows:
\begin{itemize}
    \item We propose a fully 3D-based fidelity evaluation metric for textured 3D shapes, without the need for rendering, which avoids the uncertainty introduced by viewpoint changes.
    \item We design a 3D-geometry-based architecture that jointly encodes geometric and color features of 3D shapes, enabling accurate fidelity representation and evaluation.
    \item To train and validate our metric, we design a human-annotated \ds{} dataset. We create distortions with real 3D reconstruction and generation techniques to minimize domain bias and display each model from diverse viewpoints to reduce viewpoint ambiguity in the annotation process.
\end{itemize}

Extensive experiments demonstrate that our proposed method achieves superior performance on multiple real-world reconstruction and generation datasets, significantly outperforming traditional 2D-rendering-based and geometry-only comparison methods.

\section{Related Work}

\textbf{Metrics for 3D Generation and Reconstruction.}
Evaluating the fidelity of 3D shapes is crucial for assessing the performance of reconstruction and generation methods such as \cite{borgefors1984CD, luan2021pc, zhai2023language, luan2023high, gong2022self,  zhao2025pp, luan2024divide, luan2025scalable, gong2023progressive, wu2024fsc, song2022pref, zhang2021learning, yang2025fast3r, huang2025spar3d, cui2024lam3d, reka2025multi, xiang2024structured, liu2024isotropic3d, lee2024text,liu2021nech, liu2022depth}. Traditional metrics such as Chamfer Distance (CD) \cite{borgefors1984CD} and Earth Mover's Distance (EMD) \cite{rubner1998metric} are widely used in the 3D reconstruction approaches. CD measures the average closest point distance between two point sets, and EMD provides a more accurate assessment by computing the minimal cost of transforming one distribution into another. Learning metrics that are based on CD are also proposed, including Learnable Chamfer Distance (LCD) \cite{huang2024learnable}, Density-aware Chamfer Distance (DCD) \cite{wu2021density}, and Hyperbolic Chamfer Distance (HyperCD) \cite{lin2023hyperbolic}. 
Despite these advancements, many of these metrics focus solely on geometric discrepancies and often overlook the perceptual aspects of texture and color, which are essential for human-aligned evaluations. In 3D shape generation, evaluation metrics play a crucial role in quantifying the similarity between generated results and ground-truth shapes. Metrics such as Fréchet Inception Distance (FID) have been adapted from 2D image synthesis to 3D domains by projecting shapes into 2D renderings~\cite{achlioptas2018learning}. Such projection-based evaluations suffer from viewpoint dependency and may fail to reflect the structural and textural fidelity of the original 3D shapes. Our work addresses this gap by proposing a metric that directly operates on textured 3D shapes without relying on rendering or projection.

\begin{figure*}[tb]
  \centering
   \includegraphics[width=1\linewidth]{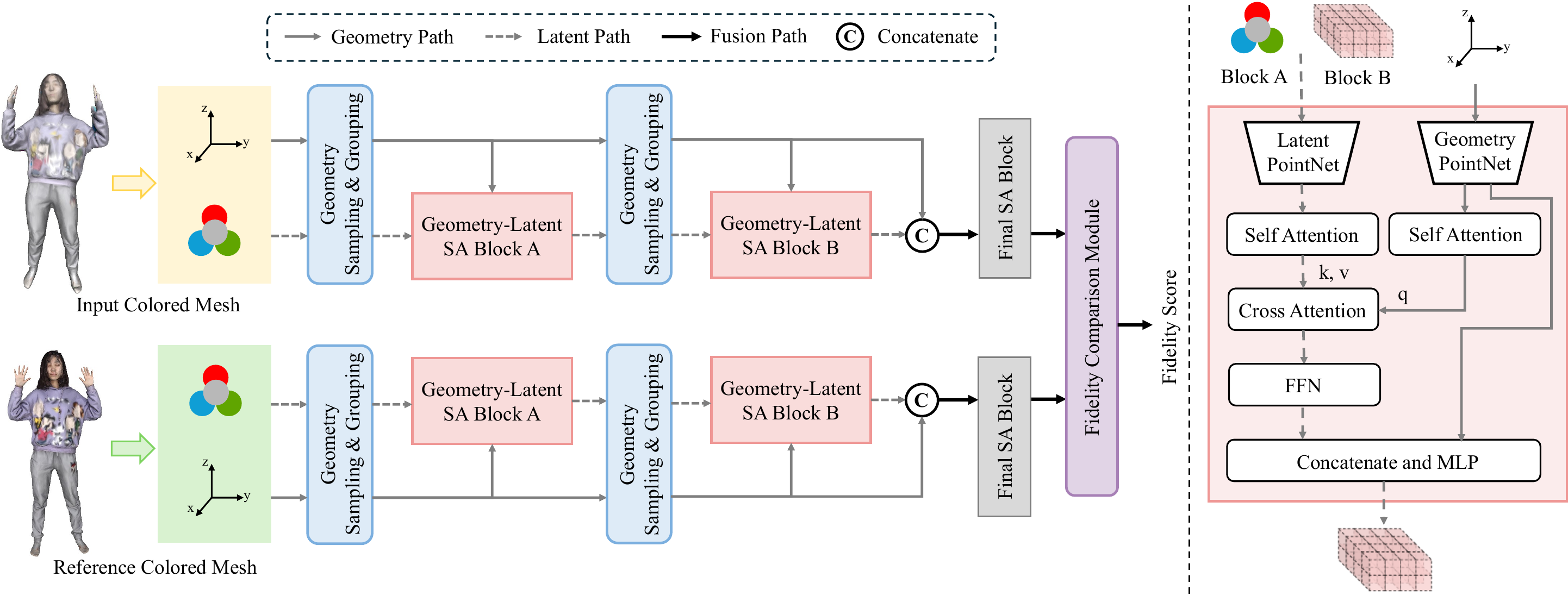}
   \caption{(a) Overview of the TGE pipeline. Given a pair of textured 3D meshes (input and reference), we extract hierarchical geometry and color features using a PointNet++-style pipeline. A novel Latent-Geometry Set Abstraction (LG-SA) block is introduced to jointly fuse geometry and color information at each level. The resulting global features from both meshes are compared by a shared MLP to predict a scalar fidelity score. This design allows perceptual fidelity evaluation without any rendering. (b) Illustration of the Latent-Geometry Set Abstraction (LG-SA) block. The module extracts geometry and appearance features via parallel self-attention modules and fuses them using a cross-attention mechanism. Geometry features serve as the query to attend to latent color features.
    }
   \label{fig:pipeline}
\end{figure*}

\textbf{Recent Advances in 3D Shape Evaluation Metrics.}
To bridge the gap between computational metrics and human perception, recent research has introduced evaluation methods that incorporate both geometric and appearance features. Graphics-LPIPS \cite{nehme2023textured} extends the Learned Perceptual Image Patch Similarity (LPIPS) metric to 3D graphics by assessing the perceptual similarity of rendered images using an LPIPS \cite{zhang2018unreasonable}-based approach. SJTU-TMQA \cite{cui2024sjtu} and TSMD \cite{yang2023tsmd} are datasets designed for textured mesh quality assessment, providing benchmarks for evaluating the visual quality of 3D models. CMDM \cite{nehme2021cmdm} introduces a full-reference metric that combines curvature-based geometric features with color information to assess the quality of colored meshes. While these methods represent significant strides toward perceptually aligned evaluations, they often rely on 2D renderings of 3D models, making them susceptible to variations in viewpoint and lighting conditions. Furthermore, the dependence on synthetic distortions in some datasets may not accurately reflect the complexities encountered in real-world scenarios, highlighting the need for evaluation metrics that operate directly on 3D data and are trained on real-world distortions.

\section{Methods}

\subsection{Problem Formulation}
We aim to design a human-aligned fidelity metric that quantifies the perceptual similarity between an input textured 3D shape and a reference shape. Given an input 3D mesh $\hat{m}$ and its corresponding ground-truth mesh $m$, our metric $F(\hat{m}, m; \theta)$ with learnable parameters $\theta$ outputs a scalar fidelity score $\hat{s} \in \mathbb{R}$:
\begin{equation}
  \hat{s} = F(\hat{m}, m; \theta),
  \label{eq:task1}
\end{equation}
where a higher score indicates higher perceptual similarity. Each mesh $m = \{(v_i, c_i)\}_{i=1}^N$ consists of $N$ vertices, with $v_i \in \mathbb{R}^3$ as the 3D coordinate and $c_i \in \mathbb{R}^3$ as the RGB vertex color. During training, we minimize the discrepancy between predicted scores and human-annotated ground-truth scores $s$:
\begin{equation}
  \min_\theta \mathcal{L}(\hat{s}, s),
  \label{eq:task2}
\end{equation}
where $\mathcal{L}$ is a hybrid loss function designed to align with both absolute and relative score accuracy (see Section~\ref{sec:training}).

\subsection{Pipeline Overview}
\label{sec:pipeline}
To align with human perceptual judgment and overcome the limitations of rendering-based evaluations, we propose the \textit{Textured Geometry Evaluation} (TGE) framework. As shown in \cref{fig:pipeline}(a), our model directly processes 3D textured meshes without rendering, and extracts features through a two-stream design that jointly models geometry and color. Our framework is designed based on PointNet++ \cite{qi2017pointnet++}. First, given a pair of 3D meshes $(\hat{m}, m)$, where $\hat{m}$ is the input mesh to be evaluated and $m$ is the reference, we first sample points from each mesh and extract local geometric features from vertex coordinates $\{v_i\}$ using sampling and grouping from \cite{qi2017pointnet++}. For each level of the Set Abstraction block, we design a novel \textit{Latent-Geometry Set Abstraction} (LG-SA) block, in which the new LG-SA block uses both color and geometry information in the latent space, while the original SA block only considers the 3D geometry information. This design is motivated by our observation that human perception relies on both shape and appearance cues. By explicitly fusing these two streams, we aim to bridge the gap between geometric alignment and perceptual realism. In Sec. \nameref{sec:glsa}, we will elaborate on the detailed design of our LG-SA block. After 2 layers of sampling \& grouping along with LG-SA block, we use the original final SA block in \cite{qi2017pointnet++} to get the encoded feature of the input mesh. For the referenced mesh, we use the same network to extract the 3D geometry-color feature.

After feature extraction, we obtain extracted feature $\mathbf{f}_{\text{input}}$ and $\mathbf{f}_{\text{ref}}$ for the distorted and reference meshes, respectively. These features are passed into a fidelity comparison module $\mathcal{C}$:
\begin{equation}
s = \mathcal{C}(\mathbf{f}_{\text{input}}, \mathbf{f}_{\text{ref}}),
\end{equation}
where $\mathcal{C}$ is a multi-layer perceptron that predicts the final scalar fidelity score $s$. This score reflects the perceptual alignment between the input and reference shapes, taking into account both geometry and texture in a rendering-free manner. In summary, our pure 3D-based structure is carefully designed to address the 2 major challenges in 3D textured mesh fidelity evaluation: avoiding viewpoint sensitivity caused by rendering, and jointly utilizing texture and 3D geometry as an essential perceptual cue.

\subsection{Latent-Geometry Set Abstraction (LG-SA) Block}
\label{sec:glsa}

To effectively combine geometry and appearance features, we design the \textit{Latent-Geometry Set Abstraction (LG-SA)} block, which is a core component of our network. This block extends the original Set Abstraction (SA) module in PointNet++~\cite{qi2017pointnet++} by introducing color-aware fusion in the latent space, thus enabling perceptually informed encoding of textured 3D shapes.

Each LG-SA block receives a set of 3D point coordinates $\{v_i\}_{i=1}^N$ and their corresponding RGB vertex colors $\{c_i\}_{i=1}^N$ as input. These inputs are processed in the geometry path and the latent feature path. For geometry path, we first use the standard farthest point sampling (FPS) and ball query~\cite{qi2017pointnet++} to select a subset of points and their local neighborhoods. Each neighborhood is encoded using a PointNet module to extract geometry features $\mathbf{g}_i \in \mathbb{R}^3$ for each sampled point. To further enhance local spatial relations, we apply a self-attention mechanism:
\begin{equation}
\mathbf{g}_i' = \text{Attn}(\mathbf{g}_i, \mathbf{g}_i, \mathbf{g}_i).
\label{eq:selfattn_geom}
\end{equation}

Similarly, we encode latent feature $\mathbf{l}_i \in \mathbb{R}^{d_l}$ using another PointNet followed by a self-attention module to yield latent features $\mathbf{l}_i' \in \mathbb{R}^{d_l}$. The color-based self-attention is defined as:
\begin{equation}
\mathbf{l}_i' = \text{Attn}(\mathbf{l}_i, \mathbf{l}_i, \mathbf{l}_i).
\end{equation}
Here, $d_l$ is the latent feature dimension. For the first LG-SA block, the input feature is just the RGB color and $d_l=3$. For the second LG-SA block, the input feature is the output feature of the first LG-SA block, where $d_l=256$.

After obtaining the refined features $\mathbf{g}_i'$ and $\mathbf{l}_i'$, we perform cross-attention to allow geometric structure to attend to appearance cues:
\begin{equation}
\mathbf{f}_i = \text{CrossAttn}(\mathbf{g}_i', \mathbf{l}_i', \mathbf{l}_i') + \text{FFN}(\mathbf{l}_i'),
\label{eq:glsa_cross}
\end{equation}
where $\mathbf{g}_i'$ is used as query $\mathbf{q}_i$, and $\mathbf{l}_i'$ as key and value. This cross-attention allows each point’s geometric understanding to be modulated by its corresponding appearance context, important for capturing visual distortions. The final fused feature $\mathbf{f}_i$ is then passed through an MLP and aggregated via max pooling to obtain a global shape-level descriptor. This design is inherently asymmetrical: geometry guides the attention, and appearance complements it. This design also reflects how human perception evaluates 3D shapes: structural integrity is the foundation, while texture is the most crucial content of the fidelity evaluation.

Importantly, this LG-SA block avoids dependency on mesh connectivity or 2D projection. By integrating local reasoning (via self-attention) and cross-modal modulation (via cross-attention), the LG-SA block captures shape and appearance interaction in a way that aligns with perceptual fidelity.

\subsection{Training Strategy}
\label{sec:training}
Our training strategy is fully supervised and end-to-end, leveraging human-labeled fidelity scores from a dataset that includes real-world distortions. To ensure that the predicted fidelity score $\hat{s}$ aligns not only numerically but also highly correlated with human labels $s$, we optimize a hybrid loss function:
\begin{equation}
\mathcal{L} = \lambda_{\text{smooth}} \mathcal{L}_{\text{smooth}} + \lambda_{\text{plcc}} \mathcal{L}_{\text{plcc}} + \lambda_{\text{srocc}} \mathcal{L}_{\text{srocc}},
\end{equation}
where The weights $\lambda_{\text{smooth}}, \lambda_{\text{plcc}}, \lambda_{\text{srocc}}$ are weight hyperparameters, and:

\textbf{Smooth L1 loss}
\begin{equation}
\mathcal{L}_{\text{smooth}} = \frac{1}{N}\sum_{i=1}^N 
\begin{cases} 
\frac{1}{2} (\hat{s}_i - s_i)^2, & \text{if } |\hat{s}_i - s_i| < 1 \\
|\hat{s}_i - s_i| - \frac{1}{2}, & \text{otherwise}
\end{cases}
\end{equation}
This loss penalizes numerical discrepancies. Here, $N$ is the number of data sample, $\hat{s}_i$ and $s_i$ are the predicted and ground-truth scores for the $i$-th sample.

\textbf{Pearson’s correlation loss (PLCC Loss)}
\begin{equation}
\mathcal{L}_{\text{plcc}} = 1 - \frac{\sum_{i=1}^N (\hat{s}_i - \bar{\hat{s}})(s_i - \bar{s})}{\sqrt{\sum_{i=1}^N (\hat{s}_i - \bar{\hat{s}})^2} \sqrt{\sum_{i=1}^N (s_i - \bar{s})^2}},
\end{equation}
This loss maximizes the linear correlation between predictions and human labels. $\bar{\hat{s}}$ and $\bar{s}$ are batch-wise means.

\textbf{Spearman’s rank-order correlation loss (SROCC Loss)}
\begin{equation}
\mathcal{L}_{\text{srocc}} = 1 - \frac{6 \sum_{i=1}^N (R(\hat{s}_i) - R(s_i))^2}{N(N^2 - 1)}
\end{equation}
This term promotes correct ranking order, which is crucial when judging relative fidelity. $R(\cdot)$ is the soft-ranking operator as in \cite{blondel2020softrank}, making the function differentiable.

\begin{figure*}[tb]
  \centering
   \includegraphics[width=0.7\linewidth]{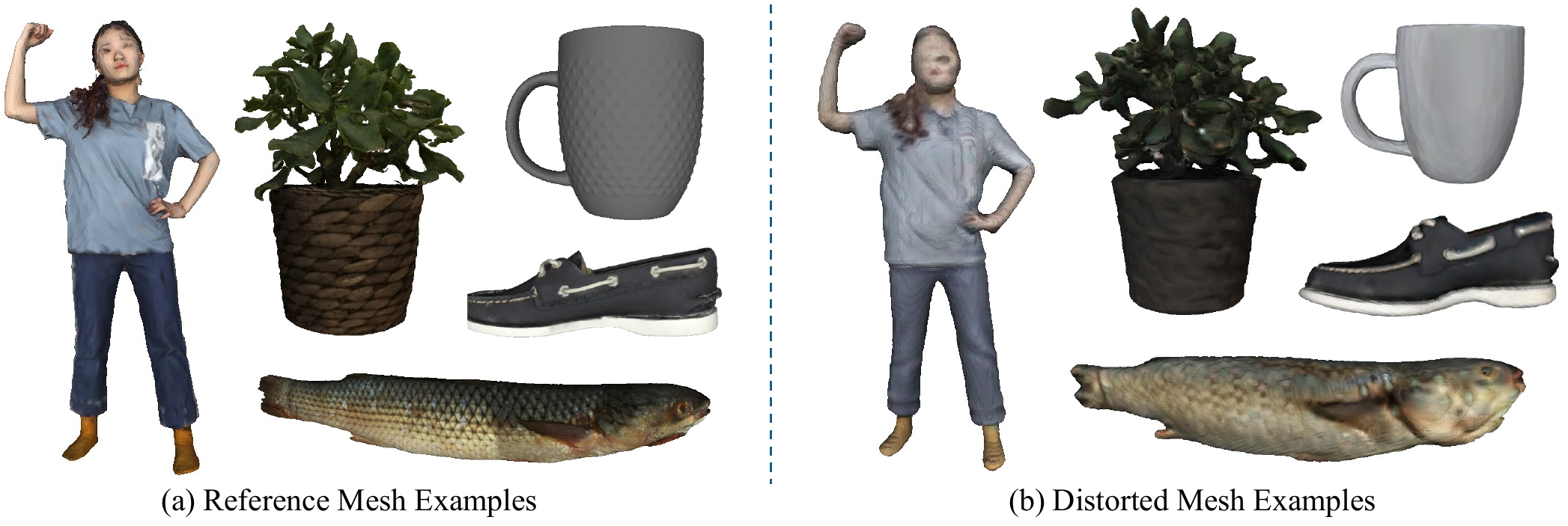}
   \caption{Some examples in our \ds{} dataset: (a) Referenced meshes. (b) Distorted meshes reconstructed or generated from real-world methods.}
   \label{fig:data}
\end{figure*}

\section{\ds{} Dataset}

To evaluate the perceptual fidelity of textured 3D shapes in realistic settings, we construct a new human-annotated dataset, denoted as \ds{}. Unlike prior benchmarks that rely on synthetic distortions or rendering-based supervision, our dataset is grounded in real-world 3D reconstruction and generation pipelines, annotated through rigorous perceptual protocols. The construction process involves three stages: distortion collection, perceptual scoring, and reliability verification.

\textbf{Real-World distortion collection.} 
We begin by selecting a diverse set of object categories and corresponding reference meshes from multiple 3D datasets, including Function4D~\cite{yu2021function4d}, ARC3D~\cite{arc3d}, RenderBot~\cite{downs2022google}, Sketchfab~\cite{sketchfab}, and CGTrader~\cite{cgtrader}. These objects serve as high-fidelity references for evaluating distortion quality.
For each object, we generate distorted meshes using a set of state-of-the-art reconstruction and generation methods, spanning from neural implicit reconstruction~\cite{xiu2022icon,xiu2023econ,saito2020pifuhd} to recent text-guided generative models~\cite{tang2023dreamgaussian,wang2024crm,xu2024instantmesh,TripoSR2024}. For text-based models, image captioning tools are first used to generate prompts, which are then fed into text-to-3D models to generate shape variants. This procedure ensures that the distortions reflect actual artifacts encountered in deployed systems, rather than artificial degradations.

\textbf{Perceptual annotation protocol.}
To ensure perceptual consistency in scoring, we adopt a pairwise comparison protocol following the Swiss tournament system used in ~\cite{luan2024spectrum}. For each object-material combination, subjects compare the distorted meshes over 6 rounds of head-to-head matchups, where each mesh competes against dynamically selected peers. Scores are assigned based on win count, ranging from 0 (loses all rounds) to 6 (wins all rounds), and subsequently normalized to the $[0,1]$ range. In total, we collect ratings from 180 unique human subjects across all object-material pairs. Each pair is evaluated by around 20 participants to reduce individual bias. In total, 1,696 annotations are obtained. This pairwise approach enforces score differentiation and prevents clustering artifacts common in absolute scoring settings.

\textbf{Outlier handling and confidence estimation.}
To enhance annotation reliability, we detect and exclude outlier scores using the interquartile range (IQR) method~\cite{dekking2005modern}. Any score deviating more than $1.5\times$IQR from the 25\textsuperscript{th} or 75\textsuperscript{th} percentile is removed. This simple but robust strategy eliminates $7.0\%$ of extreme annotations. We further quantify the consistency of annotations by computing the $95\%$ confidence interval of each object's mean score using:$\sigma_{\bar{x}} = \frac{z_{0.95} \cdot \sigma}{\sqrt{N}}$,
where $\sigma$ is the sample standard deviation, $N$ is the number of subjects per object, and $z_{0.95} \approx 1.96$ is the z-score for a $95\%$ confidence level. Our statistic shows that the IOR process reduces the overall $95\%$ confidence interval from 0.0823 to 0.067.

\textbf{Evaluation metrics.}
To evaluate how well automated metrics align with human perception, we adopt three standard correlation measures: Pearson Linear Correlation Coefficient (PLCC)~\cite{pearson1920plcc}, Spearman Rank Correlation Coefficient (SROCC)~\cite{spearman1910srocc}, and Kendall Rank Correlation Coefficient (KROCC)~\cite{kendall1946advanced}. These metrics respectively capture linear relationships, rank consistency, and ordinal agreement between predicted scores and human annotations. Each measure outputs values in the range $[-1, 1]$, with higher values indicating stronger perceptual alignment.
In \cref{fig:data} we show a few examples of the groundtruth and distortions in the dataset.

\begin{table*}
\centering
\resizebox{0.8\linewidth}{!}{
\begin{tabular}{l|ccccccccccccc}
\toprule
Method & 1 & 2 & 3 & 4 & 5 & 6 & 7 & 8 & 9 & 10 & 11 & Average$\uparrow$ & Std$\downarrow$\\
\midrule
CD      & 0.6667 & 0.4959 & 0.7288 & 0.8090 & 0.7987 & 0.4820 & 0.9003 & 0.8191 & 0.5386 & -0.0538 & 0.8046 & 0.6355 & 0.2570\\
IoU     & 0.2295 & 0.6756 & 0.5788 & 0.4970 & 0.4865 & 0.0943 & 0.8734 & 0.6475 & 0.5247 &  0.5705 & \textbf{0.8427} &0.5473 & 0.2205\\
F-score & 0.1738 & 0.0081 & 0.5147 & 0.7430 & 0.6970 & 0.4409 & 0.7770 & 0.7861 & 0.1562 &  0.1058 & 0.7429  &0.4678 &0.2907\\
P2S     & 0.7736 & 0.2731 & 0.6993 & \textbf{0.8334} & \textbf{0.8263} & 0.4206 & 0.8461 & 0.7987 & 0.4821 & -0.0363 & 0.7252 &0.6038 &0.2729  \\
ND      & 0.2091 & 0.4168 & 0.3633 & 0.7252 & 0.5080 & 0.0341 & 0.2543 & 0.2740 & -0.3038 & -0.0663 & 0.0335  &0.2031 &0.2803\\
UHD     & 0.7523 & 0.5299 & 0.4460 & 0.2933 & 0.3440 & 0.4479 & 0.7956 & 0.6987 & 0.5159 & -0.2840 & 0.7359 &0.4796 & 0.2908\\
G-LPIPS  &0.7280 & 0.7399 & 0.7089 & 0.6173 & 0.8128 & 0.5663 & \textbf{0.9151}& 0.6291 & \textbf{0.8923} & 0.5205 & 0.3584 &0.6808 & 0.1575\\
\midrule
Ours    & \textbf{0.8358} & \textbf{0.8387} & \textbf{0.7311} & 0.7951 & 0.8024 & \textbf{0.8132} & 0.6585 & \textbf{0.8737} & 0.8812 & \textbf{0.6424} & 0.8042 & \textbf{0.7887} &\textbf{0.0758}\\
\bottomrule
\end{tabular}
}
\caption{Per-object evaluation results across 11-fold object-level cross-validation using PLCC. Each object is used once as the held-out test set to assess generalization. Our method consistently achieves the highest overall performance and the lowest standard deviation across all metrics, demonstrating superior perceptual alignment and generalizability. The range of PLCC is [-1, 1], and higher values indicate stronger correlations. \textbf{Bold} number means the best.}
\label{tab:plcc}
\end{table*}

\begin{table*}
\centering
\resizebox{0.8\linewidth}{!}{
\begin{tabular}{l|ccccccccccccc}
\toprule
Metric & 1 & 2 & 3 & 4 & 5 & 6 & 7 & 8 & 9 & 10 & 11 & Average$\uparrow$ & Std$\downarrow$ \\
\midrule
CD      & 0.6167 & 0.5523 & 0.6071 & 0.5238 & 0.7381 & 0.7381 & 0.8286 & 0.7143 & 0.7381 & 0.0238 & 0.7866  &0.6243 &0.2113 \\
IoU     & 0.3167 & 0.7782 & 0.7500 & 0.2857 & 0.2169 & 0.0476 & \textbf{0.9429} & 0.5952 & 0.5476 & 0.5714 & \textbf{0.9624 }&0.5468 & 0.2871 \\
F-score & 0.0833 & 0.0187 & 0.5988 & 0.2857 & -0.1325 & 0.3810 & 0.2571 & 0.6667 & 0.0238 & 0.1429 & 0.7280  & 0.2776 & 0.2737 \\
P2S     & 0.6667 & 0.5523 & 0.5714 & 0.5714 & -0.0361 & 0.1667 & 0.9429 & 0.7381 & 0.7381 & -0.2381 & 0.7364 &0.4918 & 0.3502 \\
ND      & 0.4333 & 0.5439 & 0.4643 & 0.7143 & 0.3253 & 0.0714 & 0.2571 & 0.0714 & 0.0952 & -0.0952 & 0.3849  & 0.2969 &0.2312 \\
UHD     & 0.5833 & 0.5105 & 0.7143 & 0.1190 & 0.3494 & 0.3810 & 0.7714 & 0.5238 & 0.7857 & -0.1429 & 0.7029 &0.4817  & 0.2766\\
G-LPIPS  & 0.6333 & 0.7667 & 0.7748 & 0.2410 & 0.6506 & 0.4458 & 0.9276 & 0.7274 & 0.6752 & 0.5988 & 0.3530  & 0.6177 & 0.1911\\
\midrule
Ours    & \textbf{0.8833} & \textbf{0.8167} & \textbf{0.8469} & \textbf{0.8193} & \textbf{0.8470} & \textbf{0.8193} & 0.7247 & \textbf{0.9092} & \textbf{0.8593} & \textbf{0.6587} & 0.9160 &\textbf{0.8273} & 
\textbf{0.0731} \\
\bottomrule
\end{tabular}
}
\caption{Per-object evaluation results across 11-fold object-level cross-validation using SROCC. The range of SROCC is [-1, 1]. Higher values indicate stronger correlations. \textbf{Bold} number means the best.}
\label{tab:srocc}
\end{table*}

\section{Experiments}
\subsection{Implementation Details}
We follow the architectural design of the multi-scale grouping (MSG) variant of PointNet++~\cite{qi2017pointnet++}. Our fidelity module comprises a three-layer multilayer perceptron (MLP) with hidden dimensions of 1024, 512, and 256, respectively. For the attention module, each head is followed by a feed-forward network (FFN) consisting of two linear layers: the first expands the embedding dimension \(d_{\text{emb}}\) to \(4d_{\text{emb}}\), followed by a ReLU activation, and the second projects it back to \(d_{\text{emb}}\). Here, \(d_{\text{emb}}\) denotes the output channel of the last MLP layer in the PointNet-based encoder.
During training, we minimize a weighted sum of three loss terms: smooth loss, PLCC loss, and SROCC loss. The corresponding weights \(\lambda_{\text{smooth}}\), \(\lambda_{\text{plcc}}\), and \(\lambda_{\text{srocc}}\) are set to 1, 0.2, and 0.2, respectively. The model is trained on a single NVIDIA RTX A6000 GPU using the AdamW optimizer~\cite{loshchilov2017decoupled}, with a learning rate of \(1 \times 10^{-3}\) and a weight decay of \(1 \times 10^{-4}\). We use a batch size of 3. The implementation is based on PyTorch~\cite{paszke2019pytorch}.

\subsection{Experiment Results}
\textbf{Human alignment evaluation and generalization analysis.}
To rigorously assess the generalization ability of our proposed metric, we adopt an 11-fold object-level cross-validation strategy. Specifically, we treat each of the 11 objects in our dataset as the held-out test set in turn, while training the model on the remaining 10 categories. This object-wise split ensures that the model is always tested on \textit{unseen objects}, thereby validating the metric's performance along with generalizability on shapes. For each fold, we compute three commonly used perceptual correlation metrics, PLCC, SROCC, and KROCC, between the predicted scores and human annotations on the testing object. We report both the per-object results and the average performance across all folds, along with the standard deviation to reflect score consistency. The comparison methods include: Chamfer Distance (CD) \cite{borgefors1984CD}, Intersection of Union (IoU) \cite{henderson2018iou}, F-score \cite{wang2018fscore}, Plane-to-Surface (P2S), Normal Difference, Unidirectional Haudorff Distance (UHD) \cite{wu2020uhd}, and Graphic-LPIPS (G-LPIPS) \cite{nehme2023textured}.

As shown in \cref{tab:plcc,tab:srocc,tab:krocc}, our method achieves the highest mean performance across all three metrics, while also maintaining the lowest standard deviations. This demonstrates not only superior prediction quality but also strong reliability across different shape types. By contrast, classical geometry-based metrics such as Chamfer Distance (CD) and IoU yield significantly higher variance and lower overall correlation with human ratings. Learning-based metrics like G-LPIPS show higher accuracy but also suffer from performance fluctuations across folds. These results highlight the importance of integrating texture and geometry in a 3D-native manner, as implemented in our method. Our evaluation setup and consistent cross-object results collectively demonstrate that the proposed fidelity metric is generalizable, stable, and better aligned with perceptual evaluation than existing alternatives.

\begin{table*}
\centering
\resizebox{0.8\linewidth}{!}{
\begin{tabular}{l|ccccccccccccc}
\toprule
Metric & 1 & 2 & 3 & 4 & 5 & 6 & 7 & 8 & 9 & 10 & 11 & Average$\uparrow$ & Std$\downarrow$ \\
\midrule
CD      & 0.3889 & 0.3662 & 0.5238 & 0.2857 & 0.6429 & 0.5000 & 0.7333 & 0.5714 & 0.5000 & -0.0714 & 0.7043 & 0.4677 & 0.2161 \\
IoU     & 0.1667 & \textbf{0.6480} & 0.6190 & 0.2857 & 0.1482 & 0.0000 & \textbf{0.8667} & 0.4286 & 0.3571 & 0.4286 & \textbf{0.8733}  &0.4384 & 0.2746\\
F-score & 0.1111 & 0.0000 & 0.5822 & 0.2143 & -0.1482 & 0.2857 & 0.2000 & 0.5000 & 0.0000 & 0.0714 & 0.5916 &0.2189 & 0.2372 \\
P2S     & 0.4444 & 0.3662 & 0.4286 & 0.3571 & 0.0000 & 0.1429 & 0.8667 & 0.6429 & 0.5000 & -0.2143 & 0.6480  &0.3802 &0.2944\\
ND      & 0.3333 & 0.4789 & 0.2381 & 0.5714 & 0.2224 & 0.0714 & 0.2000 & 0.0714 & 0.1429 & -0.0714 & 0.2535  &0.2284 & 0.1757\\
UHD     & 0.5000 & 0.3662 & 0.5238 & 0.0714 & 0.3706 & 0.2857 & 0.6000 & 0.4286 & 0.6429 & -0.1429 & 0.5353 &0.3801 & 0.2250\\
G-LPIPS  & 0.5000 & 0.6111 & 0.5855 & 0.2224 & 0.5401 & 0.3706 & 0.8281 & 0.5669 & 0.5401 & 0.4001 & 0.2858  &0.4955 &0.1608\\
\midrule
Ours    & \textbf{0.7778} & 0.6111 & \textbf{0.6831} & \textbf{0.6671 }& \textbf{0.7715} & \textbf{0.6671} & 0.5521 & \textbf{0.7937} & \textbf{0.7715} & \textbf{0.4728} & 0.8003 &\textbf{0.6880} & \textbf{0.1033} \\
\bottomrule
\end{tabular}
}
\caption{Per-object evaluation results across 11-fold object-level cross-validation using KROCC. The range of KROCC is [-1, 1]. Higher values indicate stronger correlations. \textbf{Bold} numbers are the best.}
\label{tab:krocc}
\end{table*}

\begin{figure*}[ht]
  \centering
   \includegraphics[width=0.84\linewidth]{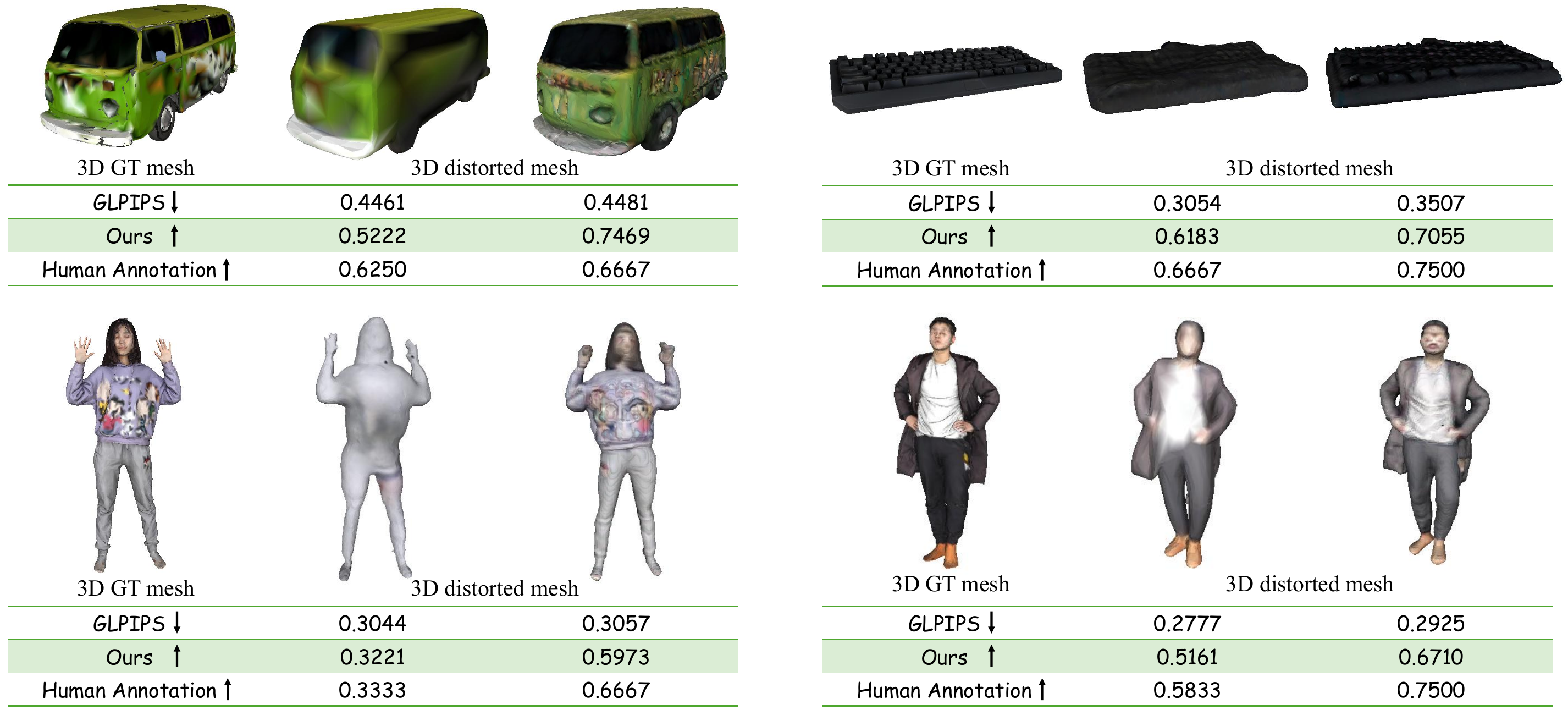}
   \caption{Visualized comparison of our metric vs. the previous metric G-LPIPS. Our metric aligns better with human annotation compared to the previous metric.}
   \label{fig:vis}
\end{figure*}

\begin{table}[ht]
\centering
\resizebox{0.8\linewidth}{!}{
\begin{tabular}{l|ccc}
\toprule
Metric & PLCC $\uparrow$ & SROCC $\uparrow$ & KROCC $\uparrow$ \\
\midrule
w/o Attention \& Latent & 0.6180 & 0.6568 & 0.5366 \\
w/o Attention  & 0.7153 & 0.7035 & 0.5924 \\
w/o Self-attention  & 0.5589 & 0.6320 & 0.5099 \\
w/o Geometry feature & 0.6519 & 0.6464 & 0.5649 \\
\textbf{Ours} & \textbf{0.7887} & \textbf{0.8273} & \textbf{0.6880} \\
\bottomrule
\end{tabular}
}
\caption{Ablation Study: impact of modules.}
\label{tab:ab1}
\end{table}

\begin{table}[ht]
\centering
\resizebox{0.85\linewidth}{!}{
\begin{tabular}{l|ccc}
\toprule
Loss function & 
PLCC $\uparrow$ & SROCC $\uparrow$ & KROCC $\uparrow$ \\
\midrule
$\mathcal{L}_{smooth}$ & 0.7742 & 0.7456& 0.6155 \\
$+\mathcal{L}_{plcc}+\mathcal{L}_{srocc}$ & 0.7296 & 0.7383 & 0.5949 \\
$+0.2\mathcal{L}_{plcc}+0.2\mathcal{L}_{srocc}$ (Ours) & \textbf{0.7887} & \textbf{0.8273} & \textbf{0.6880} \\
\bottomrule
\end{tabular}
}
\caption{Ablation Study: impact of loss functions and loss weights.}
\label{tab:ab2}
\vspace{-10pt}
\end{table}

\begin{table}[htbp]
\centering

\resizebox{\linewidth}{!}{
\begin{tabular}{l|ccccccc|c}
\toprule

Metrics & CD & IoU & F-score & P2S & ND & UHD & G-LPIPS (Learnable) & Ours (Learnable) \\
\midrule
GFLOPs & 1.6 & 0.084 & 2.0 & 2.0 & 3.0 & 0.8 & 93.11 & 14.7\\

\bottomrule
\end{tabular}
}
\caption{Computational complexity comparison of different metrics.}
\label{tab:comp}
\end{table}

\textbf{Module necessity.} In \cref{tab:ab1}, we examine different strategies for module design. The 1st row ``w/o Attention \& Latent'' is removing both the attention mechanism and the latent branch, leaving only the geometry branch, and we simply altered the original geometry Set Abstraction module from PointNet++ from 3 to 6 channels to take color input. The result justifies our design of the LG-SA module.
The 2nd row ``w/o Attention'' would remain the Latent branch design, but only remove the attention parts. This result justifies the necessity of our Attention module.
The 3rd row ``w/o Self-Attention'' has cross attention design, but removes the self-attention part. The result shows that the self-attention modules before the cross-attention module are necessary.
The 4th row ``w/o Geometry feature'' means not concatenating the geometry feature back with the latent feature after the LG-SA module. The result indicates that even when used as the query for the cross-attention module in the geometry branch, the geometry feature would still be crucial for later encoding.
Overall, the ablation proves the necessity of all proposed components in achieving optimal perceptual evaluation. 

\textbf{Loss function impacts.} In \cref{tab:ab2}, we investigate the effect of weighting in the composite loss function. Using only the Smooth L1 loss ($\mathcal{L}_{\text{smooth}}$) yields strong results. If adding a small correlation objectives ($\mathcal{L}_{\text{plcc}}, \mathcal{L}_{\text{srocc}}$) will boost the human alignment performance.

\textbf{Computational complexity.}
We compare the computational complexity of various 3D fidelity metrics in terms of GFLOPs (billion floating point operations). The number of vertices is uniformly set to 10,000. For IoU, the resolution is set to $256\times 256\times 256$, and for 
G-LPIPS, the image size is set to $1600 \times 1600$, which are both the same as \cref{tab:plcc,tab:srocc,tab:krocc}.
Traditional geometry-based metrics such as Chamfer Distance (CD), IoU, and F-score are lightweight. In contrast, recent learning-based methods like G-LPIPS incur significantly higher computational costs, reaching 93.11 GFLOPs. Our proposed method achieves a good trade-off, requiring only 14.7 GFLOPs while achieving better human alignment in fidelity. This demonstrates that our design is not only perceptually effective but also computationally efficient, making it suitable for real-world applications.

\textbf{Visualization.}
We visualize some results comparison of our metric vs. the previous metric, G-LPIPS. For G-LPIPS, the lower the better. For our metric, the higher the better. As observed, our metric aligns better with human annotation compared to the previous metric.

\section{Conclusion}

We present \textit{Textured Geometry Evaluation} (\model{}), a human-aligned fidelity metric that evaluates textured 3D meshes directly without rendering. Unlike prior methods that rely on 2D projections or synthetic distortions, \model{} jointly encodes geometry and color to assess perceptual fidelity against a reference mesh. We construct a new human-annotated dataset featuring real-world distortions to train and validate our method. Extensive experiments demonstrate that \model{} achieves better alignment with human evaluation than previous rendering-based and geometry-only approaches.

\bibliography{aaai2026}


\end{document}